# DISTAL: Distillation and Self-Supervised Pretraining for Structure-Agnostic Materials Property Prediction

Weiran Wang[1,2], Xintong Huo[1,2], Yueying Wang[1,2], Yusi Fan[1,2], Wenyan Wang[1,2], Xin Feng[3], Ruihao Xin[4], Lan Huang[1,2], Kewei Li[1,2,#], Fengfeng Zhou[1,2,#].

1 College of Computer Science and Technology, Jilin University, Changchun 130012, China.

2 Key Laboratory of Symbolic Computation and Knowledge Engineering of Ministry of Education, Jilin University, Changchun 130012, China.

3 School of Artificial Intelligence, Jilin University of Chemical Technology, Jilin 130000, PR China.

4 School of Electrical and Control Engineering, Jilin University of Chemical Technology, Jilin 130000, PR China.

# Correspondence may be addressed to Fengfeng Zhou (FengfengZhou@gmail.com or ffzhou@jlu.edu.cn), and Kewei Li (kwb1997@gmail.com).

## Abstract

Materials property prediction remains difficult in low-data settings, where many target properties are supported by only a limited number of labeled samples. Models with the strongest predictive accuracy often depend on crystal structures, which restricts their use in early-stage screening when structural information is limited or unavailable. To address this challenge, we propose DISTAL, a dual-prior framework for structure-agnostic materials property prediction that combines self-supervised compositional pretraining with structure-aware knowledge distillation. DISTAL first learns transferable compositional representations from a large virtual composition space using 145 composition-derived descriptors. It then distills structural knowledge from a pretrained ALIGNN teacher into a composition-conditioned student. This setting allows structural priors to be used during training without requiring structural inputs at inference. By integrating explicit compositional descriptors, pretrained latent features, and distilled structural features within a unified prediction pipeline, DISTAL captures complementary signals that are difficult to recover from any single representation alone. Across 39 benchmark tasks, the best-performing multimodal configuration combines all three signals, and improves over the reference benchmark on 37 tasks. DISTAL achieves the strongest overall performance among all evaluated feature combinations. These results indicate that compositional pretraining and structural distillation provide complementary priors and offer a practical route to robust composition-only prediction in small-data materials informatics. The source code and the pre-trained models are anonymously available at: https://osf.io/eq96d/overview?view_only=451617f42f7849e08750bd1852b48980 and will be released at the official link after acceptance.

## Introduction

The discovery of functional materials increasingly depends on predictive models that can prioritize large candidate spaces before costly experiments or density functional theory calculations are performed (Merchant, et al., 2023). Machine learning and deep learning have become a central tool in materials informatics, with applications across computational and experimental datasets. However, progress remains uneven across prediction tasks. A small number of properties benefit from relatively large datasets, whereas many technologically relevant properties remain in low-data regimes. This makes robust model development difficult and limits the general use of highly expressive predictors (Gupta, et al., 2024; Gupta, et al., 2021; Huang, et al., 2024; Ward, et al., 2016).

Recent progress has been driven in part by structure-aware models, especially graph neural networks that operate on atomistic representations (Chen, et al., 2019). Crystal Graph Convolutional Neural Networks showed that crystal structures can be encoded as graphs for accurate and interpretable property prediction (Xie and Grossman, 2018). Subsequent models incorporated richer geometric information. Among them, the Atomistic Line Graph Neural Network explicitly propagates both bond and bond-angle information through coupled message passing on the bond graph and its line graph, which improves prediction across diverse atomistic tasks (Choudhary and DeCost, 2021). These studies show that structural geometry provides a strong inductive bias for materials property prediction. Yet the same dependency also creates a practical barrier. Structure-aware models require crystal structures, which may be unavailable, uncertain, or computationally expensive during early-stage materials screening (Choudhary and DeCost, 2021; Xie and Grossman, 2018).

Structure-agnostic approaches address this limitation by predicting properties from composition alone. Classical descriptor-based frameworks showed that chemically informed composition-derived attributes can support broad predictive coverage across inorganic materials (Ward, et al., 2016). More recent deep learning methods, such as Roost, further demonstrated that learnable composition-only representations can improve prediction without structural inputs (Goodall and Lee, 2020). These methods are attractive for high-throughput discovery because they preserve scalability and avoid the structural bottleneck. However, composition-only learning has an intrinsic limitation. It does not directly observe geometric information that often governs coordination-sensitive, phase-dependent, or polymorph-dependent behavior. Its predictive ceiling can therefore remain below that of strong structure-aware models (Goodall and Lee, 2020; Lee, et al., 2023).

Data scarcity adds a second challenge. Many materials properties have too few labeled samples to support stable training of expressive models from scratch. Transfer learning has therefore become an important strategy in materials informatics. Gupta et al. showed that cross-property deep transfer learning can improve prediction on small materials datasets, even in composition-based settings (Gupta, et al., 2021). Later work extended this idea to structure-aware graph neural networks and reported gains across diverse materials datasets (Gupta, et al., 2024). In parallel, recent studies have shown that self-supervised and pretraining-based strategies can strengthen structure-agnostic models by learning reusable compositional representations from unlabeled data before downstream adaptation (Huang, et al., 2024; Rahman, et al., 2025). These studies suggest that transfer learning and pretraining are both useful for low-data prediction. However, they have often been developed along separate directions. One direction emphasizes knowledge transfer across labeled property datasets, while the other focuses on representation learning within composition-only frameworks (Gupta, et al., 2024; Gupta, et al., 2021; Huang, et al., 2024). These developments also align with the emerging foundation-model view in materials discovery, where large-scale pretraining is used to learn transferable representations across downstream tasks (Pyzer-Knapp, et al., 2025).

This separation leaves an important methodological gap. Composition-only models are deployable at the earliest stages of screening, but structure-aware models benefit from richer physical priors. A practical framework should retain the inference-time simplicity of composition-only models and still absorb useful information from structure-aware learning. This need motivates the present study.

Here, we introduce DISTAL, a dual-prior framework for structure-agnostic materials property prediction through distillation and self-supervised pretraining. DISTAL combines two complementary sources of prior information within a single composition-only inference pipeline. First, it learns transferable compositional priors through self-supervised pretraining on a large virtual composition space represented by 145 composition-derived descriptors. Second, it distills structure-aware knowledge from a pretrained ALIGNN teacher into a composition-conditioned student encoder. This design allows structural priors to guide training without requiring structural inputs during inference. The final prediction pipeline integrates explicit compositional descriptors, pretrained latent compositional features, and distilled structural features. DISTAL is therefore not designed to replace structure-aware learning. Instead, it provides a practical bridge between structure-aware supervision and structure-agnostic deployment.

Using this framework, we evaluate raw compositional descriptors, pretrained compositional features, distilled structural features, and their combinations across 39 benchmark tasks. Rather than treating self-supervised compositional learning and structural transfer as competing alternatives, our results support a complementary interpretation. Compositional pretraining and structural distillation encode distinct priors, and their integration yields a more robust composition-only prediction framework for low-data materials informatics.

# Materials and Methods

## Overall Workflow

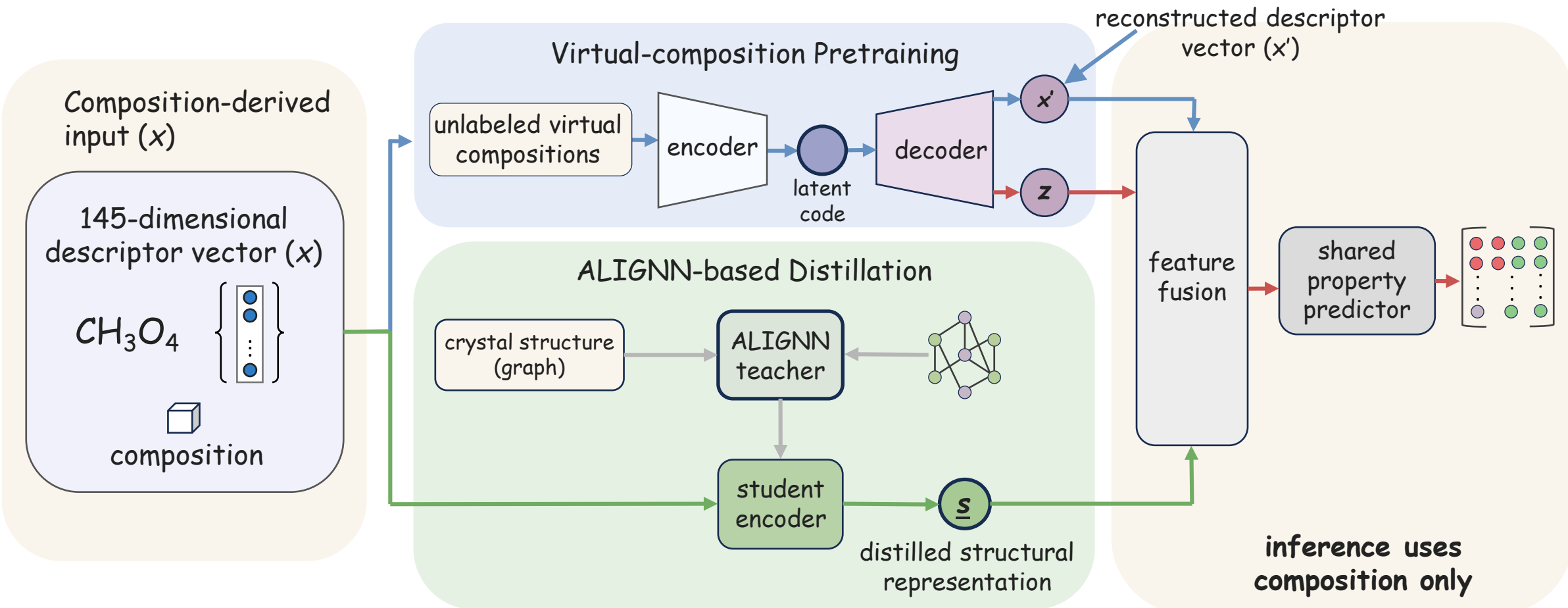

**Figure 1. Overall workflow of DISTAL for structure-agnostic materials property prediction.**

All materials were represented in a shared 145-dimensional composition-derived descriptor space, denoted as $x$. This descriptor vector served as the common input representation for the full DISTAL workflow.

As shown in Figure 1, DISTAL contains two upstream representation-learning branches and one shared downstream prediction stage. The first branch performs self-supervised compositional pretraining on a large unlabeled virtual composition set. This branch maps $x$ to a latent compositional representation, denoted as $z$, and reconstructs the input descriptor vector as $x'$ through a decoder. The reconstruction task encourages the encoder to learn compact composition-level regularities from the virtual composition space.

The second branch transfers structure-aware information from a pretrained ALIGNN teacher to a composition-conditioned student encoder. The ALIGNN teacher uses crystal-structure graphs as inputs, whereas the student encoder receives only the composition-derived descriptor vector $x$. Through this teacher-student design, the student learns a distilled structural representation, denoted as $s$. This representation carries structure-informed signals during training, but it does not require structural inputs during downstream inference.

The downstream stage evaluates $x$, $x'$, $z$, $s$, and their combinations across 39 property-prediction tasks. This workflow separates three information sources: explicit compositional descriptors, latent features learned from self-supervised compositional pretraining, and structure-informed features learned through distillation. The design allows direct assessment of whether compositional pretraining and structure-aware distillation provide complementary information for structure-agnostic materials property prediction.

## Virtual Composition Generation and Descriptor Construction

An unlabeled pretraining corpus was generated to support self-supervised compositional learning. The corpus contained 100,000 virtual compositions obtained through random formula sampling. In the implemented generator, each candidate formula contained two to four elements selected from a predefined elemental pool. Each selected element was assigned an integer stoichiometric coefficient from 1 to 4. A candidate formula was retained only when it could be parsed and featurized successfully and did not contain missing descriptor values. This virtual corpus was used to expand compositional coverage for representation learning. It was not intended to define a chemically curated materials database.

All compositions were encoded using a 145-dimensional composition-derived descriptor vector. The descriptor set included stoichiometric descriptors, Magpie elemental-property statistics, average valence-orbital features, and ion-property features. Composition parsing was performed with pymatgen (Ong, et al., 2013), and descriptor generation was implemented with matminer (Ward, et al., 2018). The Magpie descriptor family follows the general-purpose materials property prediction framework introduced by Ward et al. (Ward, et al., 2016). The same 145-dimensional representation was used throughout the study for virtual-composition pretraining, structural distillation, and downstream benchmark evaluation.

For downstream evaluation, we adopted the 39 computational prediction tasks defined in the JARVIS-based cross-property benchmark of Gupta et al. (Gupta, et al., 2021). Each task-specific dataset was converted into the common 145-dimensional descriptor space before model training. All downstream comparisons used the corresponding predefined splits.

## Self-Supervised Compositional Pretraining

Self-supervised compositional pretraining was used to learn reusable representations from the standardized virtual-composition descriptor matrix. All candidate models received the same 145-dimensional Magpie-based composition descriptor vector as input and used a fixed 72-dimensional latent space. The initial evaluation experiment included a baseline autoencoder, a denoising autoencoder (DAE) (Vincent, et al., 2010), a sparse autoencoder, a variational autoencoder (VAE) (Kingma and Welling, 2013), and a contractive-autoencoder-inspired baseline (Rifai, et al., 2011). For deterministic encoder–decoder models, the encoder compressed the input descriptor vector through a multilayer perceptron with batch normalization (Ioffe and Szegedy, 2015), LeakyReLU activation, and dropout (Srivastava, et al., 2014). The decoder then mapped the latent vector back to the original descriptor space.

The candidate models differed in how they constrained or perturbed the latent representation. The DAE added Gaussian corruption to the input descriptors. The sparse autoencoder used sparsity regularization during training. The VAE used a probabilistic latent-variable formulation. The contractive-autoencoder-inspired baseline introduced a sensitivity penalty to encourage local stability of the learned representation. All models were pretrained on the standardized virtual-composition descriptor matrix with an 8:2 train-validation split, AdamW optimization (Loshchilov and Hutter, 2017), and early stopping. No independent test subset was created at this pretraining stage because the autoencoders were not used as final predictive models. The validation split was used only to monitor pretraining and guide early stopping or representation selection. Final performance assessment was conducted exclusively on the 39 downstream property-prediction tasks.

After the generic autoencoder evaluation step, the denoising route was examined in greater detail through six variants: adaptive DAE, masked DAE, progressive DAE, context DAE, multi-scale DAE, and residual DAE. The adaptive DAE used a trainable noise parameter to learn the corruption amplitude. The masked DAE applied random feature masking. The progressive DAE adjusted the noise level across training. The context DAE used an attention-based feature-mixing block. The multi-scale DAE fused latent representations extracted under different corruption scales. The residual DAE added a residual path from the input descriptors to the reconstruction output. These variants were trained on the same standardized descriptor matrix and followed the same train–validation protocol.

The adaptive-denoising branch was then combined with additional architectural motifs to construct five hybrid backbones: ADAE-Standard, ADAE-Transformer, ADAE-GAN, ADAE-ResNet, and ADAE-CNN. These models preserved the same descriptor input and latent-space setting, but they differed in the architectural component used to enhance the encoder–decoder backbone. ADAE-Transformer introduced a projected attention branch inspired by self-attention (Vaswani, et al., 2017). ADAE-GAN added an adversarial discriminator in latent space (Goodfellow, et al., 2014). ADAE-ResNet used residual blocks in the encoder–decoder backbone (He, et al., 2016).

ADAE-CNN treated the descriptor vector as a one-dimensional signal and used convolutional encoding with transposed-convolution decoding.

Stage-wise screening of these autoencoder variants was conducted to select the final compositional encoder for downstream fusion experiments. The selected encoder generated two feature types from the same 145-dimensional descriptor vector: the latent compositional representation $z$ and the decoder-side reconstruction $x'$. These two feature types were later evaluated with the explicit descriptor vector $x$, the distilled structural representation $s$, and their combinations in the downstream prediction stage.

## ALIGNN-Guided Structural Distillation

Structural distillation was conducted to incorporate geometry-aware information into a composition-only representation. In this branch, a pretrained ALIGNN model served as the teacher (Choudhary and DeCost, 2021). ALIGNN is a graph neural network that performs message passing on both the atomistic bond graph and its line graph. This design allows the model to encode bond-level and bond-angle information within atomistic structures (Choudhary and DeCost, 2021). In the present workflow, teacher embeddings were extracted from the readout layer of a pretrained formation-energy ALIGNN model. Embeddings were generated for successfully processed structures from a JARVIS-derived corpus (Choudhary and DeCost, 2021; Choudhary, et al., 2020).

A composition-conditioned student encoder was then trained to regress the teacher embeddings from the same 145-dimensional Magpie-based composition descriptors used throughout the study. The student network mapped the descriptor input to a 256-dimensional output space through a multilayer perceptron with batch normalization, SiLU activation (Elfwing, et al., 2018), and dropout. Training was formulated as supervised regression with mean-squared-error loss. An 8:2 train-validation split and early stopping were used. The student output is denoted as $\boldsymbol{s}$ and is used as the distilled structural representation in downstream prediction. In this design, structure-derived information is transferred during training, while downstream inference requires only composition-derived descriptors.

## Downstream Feature Combinations

Four feature types were evaluated in the downstream prediction stage. The vector $\boldsymbol{x}$ denotes the original 145-dimensional Magpie-based composition descriptor vector. The vector $\boldsymbol{z}$ denotes the latent compositional representation produced by the selected pretrained encoder. The vector $x'$ denotes the decoder-side reconstruction generated by the same pretrained model in the original descriptor space. The vector $s$ denotes the distilled structural representation predicted by the composition-conditioned student encoder.

Based on these feature types, nine downstream feature combinations were examined: $x$, $x'$, $z$, $s$, $z + x$, $z + x'$, $x + s$, $x' + s$, and $z + x + s$. Feature fusion was implemented by direct vector concatenation. This design allowed the downstream evaluation to compare individual information sources with fused representations. It also tested whether pretrained compositional features and distilled structural features provided additional information beyond the original composition descriptors.

## Downstream Prediction Protocol

Downstream evaluation was performed on 39 task-specific datasets. Each dataset included a predefined training split and a fixed external test split. Cross-validation was applied only within the training split. For each task and feature combination, the downstream predictor was trained by 10-fold cross-validation on the task-specific training data. In each fold, the fold-specific training portion was used for model fitting, and the held-out fold was used for validation. The fixed external test split was not used for scaler fitting, model fitting, validation, or early-stopping decisions.

All preprocessing transformations were fitted only on training data and were then applied to the corresponding validation and test data. When target scaling was enabled, the target scaler was fitted only on the fold-specific training targets. The downstream predictor was a multilayer perceptron with three hidden layers, batch normalization, GELU activations (Hendrycks and Gimpel, 2016), dropout, and a scalar output layer. Optimization used AdamW with early stopping (Loshchilov and Hutter, 2017). Model performance was evaluated by mean absolute error.

For each task and feature combination, the final test prediction was obtained by averaging the predictions of the 10 fold-specific models on the same fixed test split. The mean absolute error of the averaged prediction was reported as the primary metric. Benchmark comparison followed the 39-task evaluation framework introduced in prior cross-property transfer-learning work (Gupta, et al., 2021).

# Results and Discussion

## Benchmark Heterogeneity and Virtual Composition Coverage

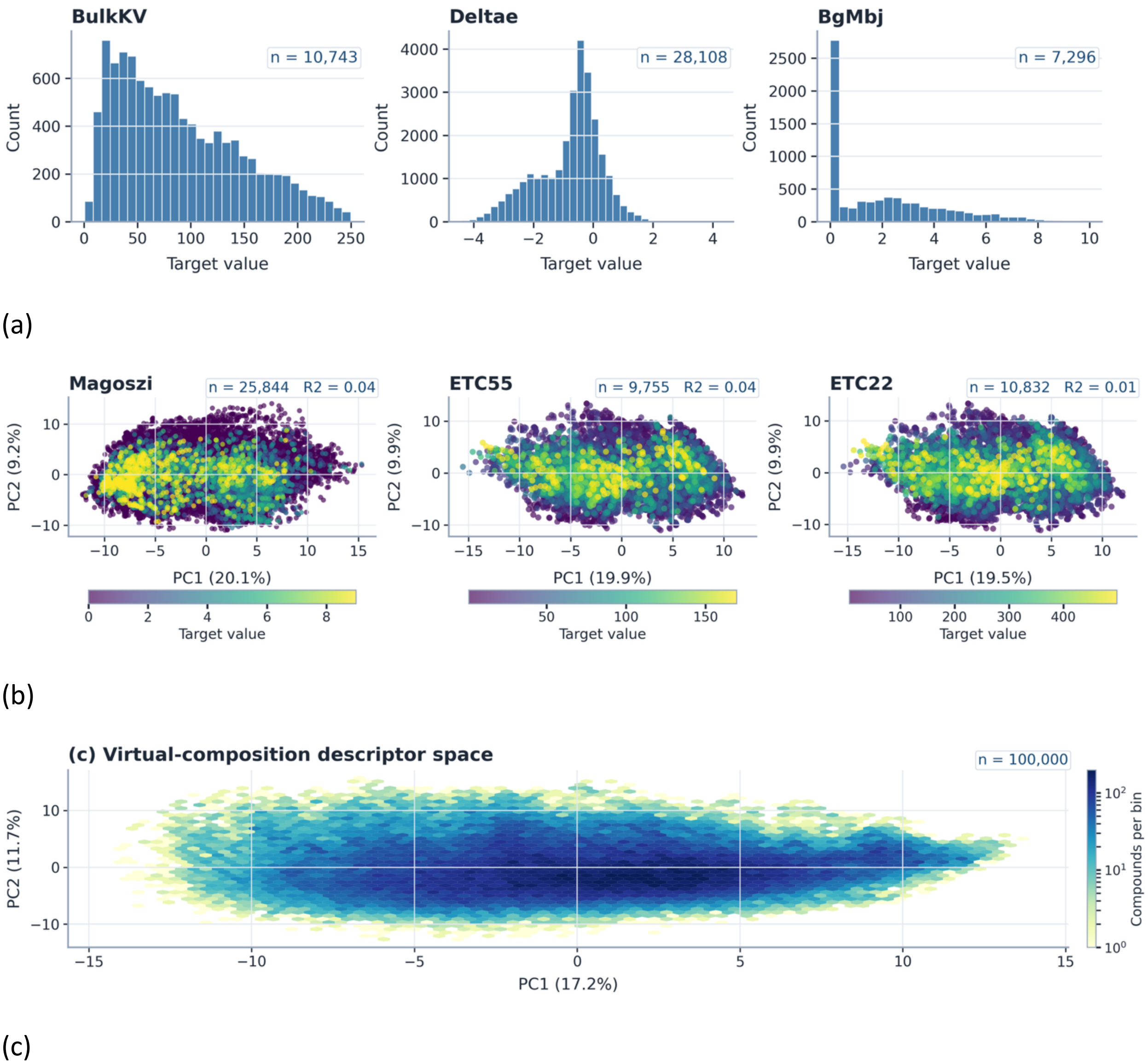

**Figure 2. Benchmark diversity and virtual-composition coverage.** The 39 downstream tasks have heterogeneous target distributions, and the virtual composition corpus provides broad unlabeled descriptor-space coverage for pretraining. (a) Representative target-property distributions, and the complete list of plots for the 39 tasks is in the Supplementary Figure S1. (b) Representative PCA projections of downstream tasks, and the complete list of the plots for the 39 tasks is in the Supplementary Figure S2. (c) PCA density map of the virtual composition corpus.

Before evaluating the learned representations, we first examined the statistical structure of the 39 downstream benchmark tasks and the coverage of the unlabeled virtual composition space used for pretraining. This analysis provides context for the downstream results. It also clarifies why a single raw descriptor representation may not be sufficient across all tasks.

The target-property distributions differed substantially across the 39 downstream tasks (Gupta, et al., 2021) (Figure 2a). Some tasks showed compact value ranges, whereas others showed broad, skewed, multimodal, or long-tailed distributions. Several properties also contained dense regions near low values with sparse high-value tails. These patterns indicate that the benchmark should not be treated as a collection of closely matched regression tasks. Instead, it covers heterogeneous prediction settings with distinct target scales and distributional shapes. This heterogeneity is important for evaluating whether learned representations can improve prediction across diverse materials-property regimes rather than only under a narrow data distribution.

We next examined the raw 145-dimensional composition-derived descriptor space by task-wise PCA projections (Figure 2b). Several datasets showed visible target-associated gradients in the first two principal components. This observation suggests that the descriptor space already retains useful property-related variation. However, these gradients were often diffuse and overlapping. The two-dimensional projections rarely showed clean local separation by target value. Thus, the raw descriptor space provides an informative starting point, but it may not organize property-relevant variation in a form that is optimal for downstream prediction. This result supports the use of representation learning as a complementary step rather than a replacement for chemically meaningful descriptors.

The PCA density map of the 100,000 virtual compositions further showed broad coverage in the Magpie-based descriptor space (Figure 2c). The virtual corpus formed a dense central region with extended peripheral coverage. This pattern is consistent with the intended role of the corpus. It was not designed as a chemically curated candidate library. Instead, it served as a large unlabeled descriptor manifold for self-supervised pretraining (Huang, et al., 2024). Such a corpus can expose the encoder to broad compositional variation before task-specific supervision is introduced.

Together, these exploratory results motivate the representation-learning design (Merchant, et al., 2023) used in DISTAL. The downstream benchmark contains strong task-level heterogeneity, and the raw descriptor space contains useful but only partially organized property information. At the same time, the virtual composition corpus provides broad unlabeled coverage for compositional pretraining. These observations support the subsequent evaluation of explicit descriptors, pretrained compositional features, distilled structural features, and their fused representations.

## Raw Composition-Derived Descriptors Establish a Strong Baseline

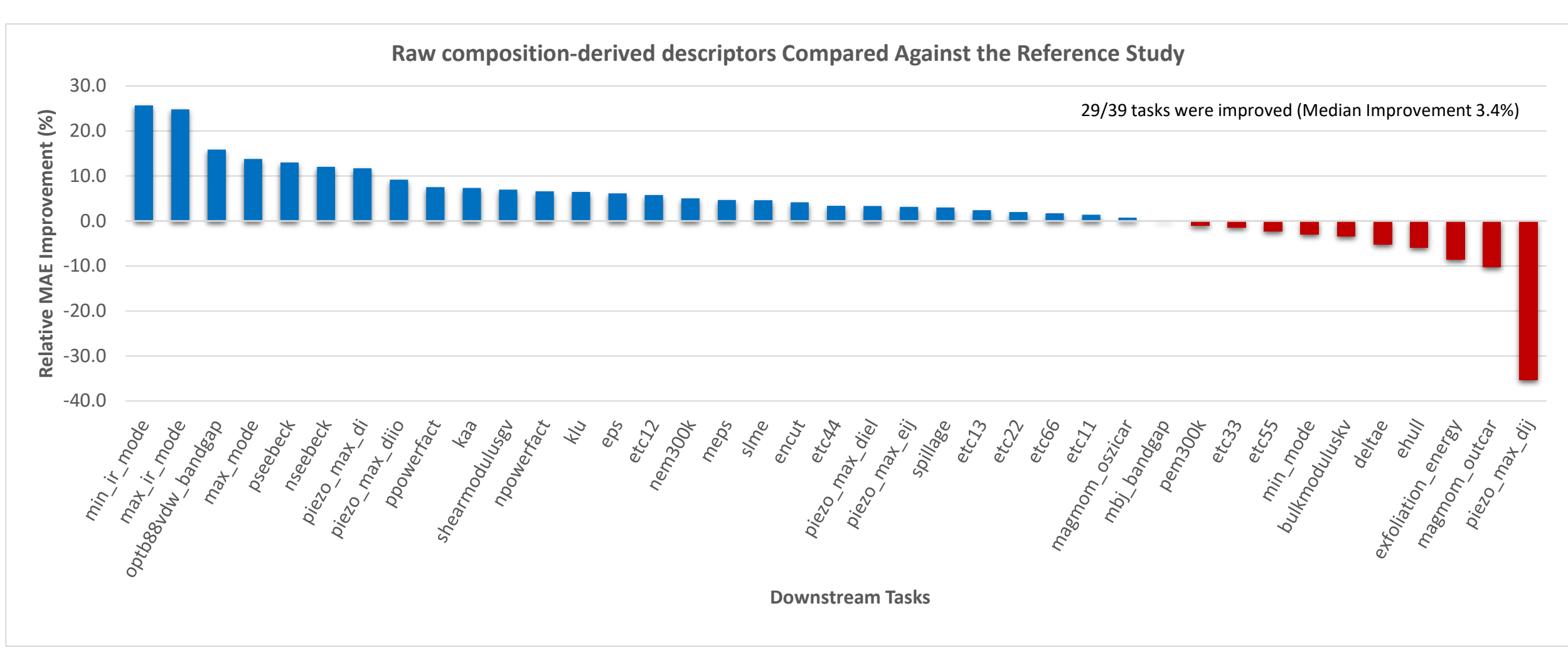

**Figure 3. Raw composition-derived descriptors as the baseline representation.** The figure compares the model trained with raw composition-derived descriptors against the reference model in terms of MAE, and their performances were denoted as $MAE_{Raw}$ and $MAE_{Ref}$. The relative improvement is calculated as $(MAE_{Ref} - MAE_{Raw})/MAE_{Ref} \times \mathbf{100}\%$. Because lower MAE indicates better predictive performance, positive values indicate that the raw descriptor model outperformed the reference model, whereas negative values indicate lower performance than the reference model. Overall, the raw descriptor baseline improved 29 of the 39 downstream tasks, with a median relative improvement of 3.4%.

We first evaluated the explicit composition-derived descriptor vector $x$ as a standalone baseline. This analysis is important because all downstream feature combinations in DISTAL are derived from, or combined with, the same descriptor space. If the raw descriptor baseline was weak, later gains from pretraining or distillation would be difficult to interpret. If $x$ already contains strong predictive signal, the learned branches should be interpreted as complementary refinements rather than replacements for composition-derived descriptors.

The raw descriptor baseline achieved competitive performance across a substantial fraction of the benchmark (Figure 3). Relative to the reference benchmark of Gupta et al. (Gupta, et al., 2021), $x$ improved the test MAE on 29 of the 39 tasks. The median relative MAE improvement was +3.4%. This result indicates that the 145-dimensional Magpie-based descriptors encode substantial transferable information across chemically diverse prediction tasks. Thus, the starting point of DISTAL is not a weak handcrafted control. It is an informative composition-only representation that remains competitive across a broad multitask benchmark.

The improvement pattern was not uniform. Several tasks showed clear gains, whereas a smaller subset remained below the reference benchmark. This task-dependent behavior agrees with the exploratory analysis above. The raw descriptor space preserves property-relevant chemical information, but this information is not always organized in a form that supports equally strong prediction across all targets. This result leaves room for self-supervised pretraining and structural distillation to reshape or augment the descriptor-derived signal before downstream prediction.

These observations define the interpretive baseline for the rest of the study. First, explicit composition-derived descriptors should be treated as a strong benchmark, not as a nominal control. Second, any benefit from learned representations should be judged against this competitive composition-only baseline. The next analysis therefore asks whether self-supervised compositional pretraining can extract latent features that further strengthen the predictive information already present in $\boldsymbol{x}$.

## Autoencoder-Derived Features Enhance Prediction

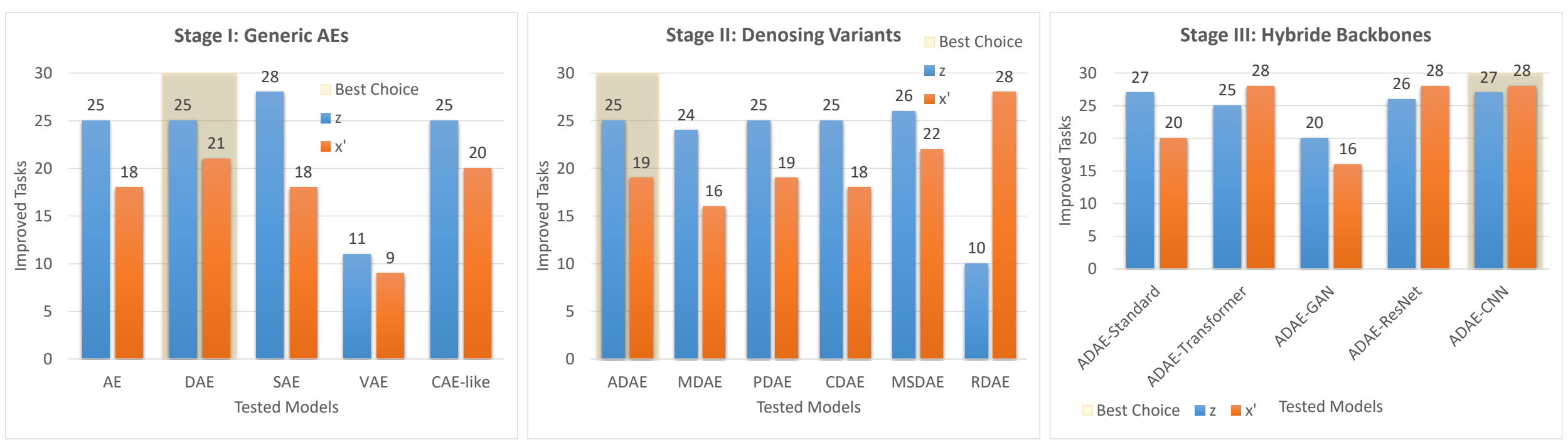

**Figure 4. Stage-wise selection of self-supervised compositional autoencoder backbones.** (a) Stage I comparison of generic autoencoder families. (b) Stage II comparison of denoising variants. (c) Stage III comparison of hybrid ADAE backbones.

We next evaluated whether self-supervised autoencoder pretraining could improve downstream predictions beyond the raw descriptor baseline. The comparison was performed in three stages, and each stage assessed both learned feature types derived from the pretrained model: the latent representation $z$ and the decoder-side reconstruction $x'$. This distinction is important because the downstream analysis does not assume that the latent space is the only useful representation. The reconstruction stream may also retain transformed descriptor information that benefits prediction.

The stage-wise comparison showed that self-supervised compositional pretraining can improve downstream predictions, but the effect was strongly dependent on the backbone and the feature type (Figure 4). No candidate was uniformly optimal for both $\boldsymbol{z}$ and $\boldsymbol{x}'$. This result indicates that latent knowledge transferability and reconstruction-side utility should be treated as related but non-identical properties of a pretrained compositional model.

In Stage I, the generic autoencoder comparison showed that deterministic autoencoder families were more competitive than the variational baseline (Figure 4a). On the latent side, the sparse autoencoder achieved the highest improvement count, with gains on 28 of the 39 tasks. On the reconstruction side, the denoising autoencoder gave the strongest result among the Stage I candidates, with improvements on 21 tasks. The denoising autoencoder was therefore retained for the next stage. This outcome suggests that input perturbation provided a useful pretraining signal for the present virtual-composition descriptor corpus. It also shows that the best latent representation and the best reconstruction stream did not necessarily arise from the same criterion.

Stage II compared six denoising variants and again showed a split between the two feature types (Figure 4b). MSDAE achieved the highest latent-side improvement count, with gains on 26 of the 39 tasks. RDAE achieved the highest reconstruction-side count, with improvements on 28 tasks, but it showed weaker latent-side prediction power. ADAE improved 25 tasks through $z$ and 19 tasks through $x'$. It was therefore retained for architectural extension because it provided a more balanced outcome across both branches. This result suggests that the denoising strategy changes not only the reconstruction behavior, but also the downstream value of the learned latent representation.

Stage III further examined hybrid ADAE backbones (Figure 4c). ADAE-CNN produced the strongest latent-side result, with improvements on 27 of the 39 tasks. The reconstruction branch reached 28-task improvements for ADAE-CNN, ADAE-ResNet, and ADAE-Transformer. Among these high-performing reconstruction-side candidates, ADAE-ResNet was retained as the final compositional backbone for the subsequent fusion experiments. This choice reflects the stage-wise selection logic of the present workflow. The final backbone was selected to balance denoising behavior, latent transferability, reconstruction-side performance, and compatibility with the later fusion setting.

Overall, the autoencoder comparison supports two conclusions. First, self-supervised pretraining on the unlabeled virtual composition space can produce feature types that improve downstream prediction across many tasks. Second, the benefit depends on the form of the learned representation. The latent branch and the reconstruction branch encode different views of the same descriptor space, and their downstream value can diverge across architectures. This finding supports the later use of both $\boldsymbol{z}$ and $\boldsymbol{x}'$ as candidate feature types rather than restricting the analysis to the latent embedding alone. It is also consistent with recent materials-informatics studies that show pretraining can improve prediction in data-limited materials settings (Huang, et al., 2024; Rahman, et al., 2025).

## Distilled Structural Features Provide the Strongest Standalone Representation

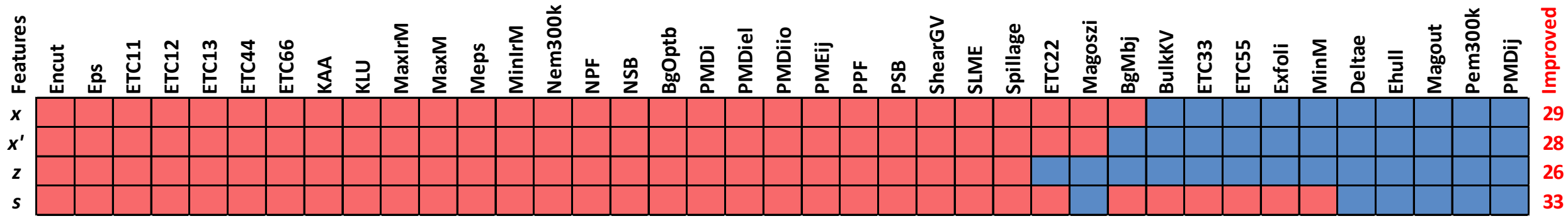

**Figure 5. Standalone benchmark comparison of four feature types.** The figure compares four standalone feature types, including $x$, $x'$, $z$, and $s$, across the 39 downstream tasks. The last column gives the number of tasks improved over the reference benchmark for each feature type. The task-wise map shows improvement status for each task and feature type. Colored cells denote improvement (red) over the reference benchmark, whereas blue cells denote no improvement. $s$ achieved the highest improvement count among the four standalone feature types.

We next evaluated the distilled structural branch $s$ as a standalone representation. This analysis isolates the effect of structure-aware knowledge transfer under the composition-only inference setting used throughout this study. The compositional branches $z$ and $x'$ were learned from the virtual descriptor manifold, whereas $s$ was obtained by training a composition-conditioned student encoder to regress embeddings from a pretrained ALIGNN teacher. The downstream behavior of $s$ therefore reflects the extent to which structure-aware information can be transferred into a model that does not require crystal structures at inference time.

The distilled structural branch was the strongest standalone feature type (Figure 5). Used alone, $s$ improved over the reference benchmark on 33 of the 39 tasks. This exceeded the raw descriptor baseline $x$ (29/39), the reconstructed compositional branch $x'$ (28/39), and the latent compositional branch $z$ (26/39) (Figure 5a). The task-wise comparison further shows that this advantage was broadly distributed across the benchmark rather than restricted to a small group of datasets (Figure 5b). These results indicate that structure-aware supervision remains informative after transfer into a composition-conditioned representation. In practical terms, part of the benefit of structure-aware learning can be retained as a distilled prior within a structure-agnostic predictor.

This finding is central to the design of DISTAL. The goal is not to reproduce a full structure-aware model under the same input conditions. Instead, the goal is to recover useful structural information when deployment is limited to composition-only inputs. From this perspective, the strong standalone performance of $s$ shows that teacher-student distillation provides an effective route for importing structural bias into a composition-only prediction pipeline. This interpretation is consistent with recent work showing that structure-aware transfer learning can strengthen materials-property prediction across diverse datasets (Gupta, et al., 2024), and with recent studies that highlight pretraining and transferred representations for data-limited materials modeling (Huang, et al., 2024; Rahman, et al., 2025).

The role of $\boldsymbol{s}$ should still be interpreted in relation to the other feature types. Its strong standalone performance does not mean that explicit compositional descriptors are unnecessary. It also does not imply that distilled structural features should replace direct structure-aware modeling when reliable crystal structures are available. Rather, the result shows that structural distillation is a practical source of transferable information under a composition-only inference constraint. This observation motivates the next analysis, where the complementarity among $x$, $z$, $x'$, and $s$ is evaluated through feature fusion.

## Feature Fusion Reveals Complementary Information

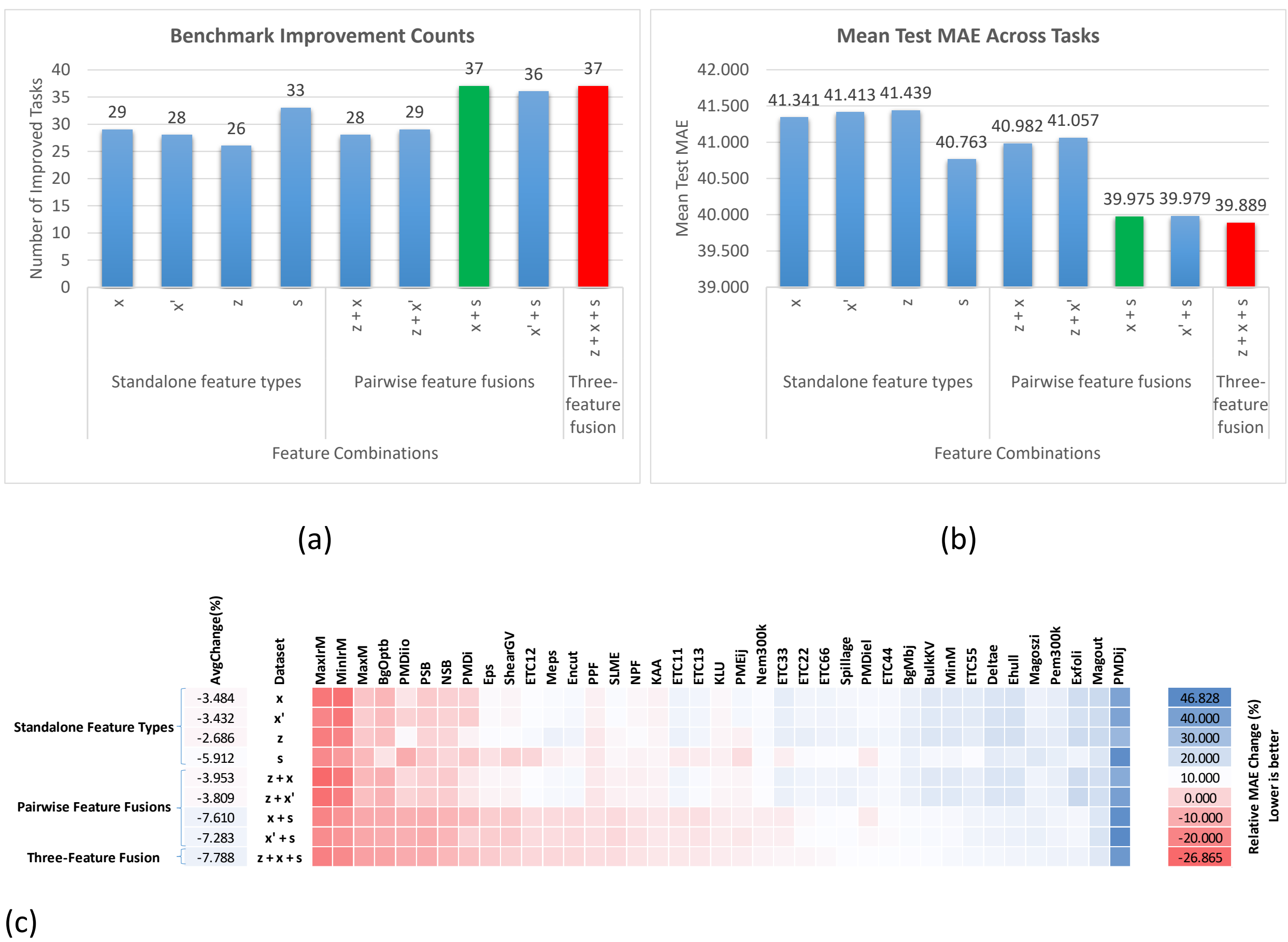

**Figure 6. Benchmark summary across nine downstream feature modes.** (a) This panel shows the number of downstream tasks improved over the reference benchmark. (b) This panel shows the mean test MAE across the 39-task benchmark. Standalone feature types are shown for $x$, $x'$, $z$, and $s$. Pairwise feature fusion modes are shown for $z + x$, $z + x'$, $x + s$, and $x' + s$. The three-feature fusion $z + x + s$ achieves the lowest mean test MAE and matched the highest improvement count. (c) Task-wise MAE improvements across nine feature modes.

The preceding analyses showed that the four feature types contribute different types of information. The explicit descriptor vector $x$ provided a strong composition-only baseline. The autoencoder-derived features $z$ and $x'$ captured self-supervised compositional structure. The distilled representation $s$ transferred structure-aware information into a composition-only inference setting. We therefore asked whether these feature types were redundant, or whether their combination could improve downstream prediction.

The comparison across nine feature combinations shows clear complementarity among the feature types (Figure 6a,b). Among the standalone features, $s$ achieved the strongest result, with improvements over the reference benchmark on 33 of 39 tasks and a mean test MAE of 40.76. However, the two-way fusion $x + s$ further improved performance, with gains on 37 of 39 tasks and a lower mean test MAE of 39.97. This result shows that the explicit compositional descriptors in $x$ are not fully captured by the distilled structural representation s. Instead, descriptor-level chemical information and transferred structural information provide distinct signals for downstream prediction. The task-wise heatmap further shows that this complementarity is not restricted to the aggregate counts: different feature combinations dominate different subsets of tasks, whereas the final $z + x + s$ representation provides one of the most consistently favorable patterns across the benchmark (Figure 6c).

The three-way fusion $z + x + s$ achieved the strongest overall result in the present comparison. It matched the highest improvement count, with gains on 37 of 39 tasks, and produced the lowest mean test MAE, 39.89. This result is notable because $z$ was not the strongest standalone feature type. Used alone, $z$ improved 26 of 39 tasks. Its added value became clearer after fusion with $x$ and $s$. This pattern suggests that the latent compositional representation captures information that is not fully preserved by the raw descriptors or by the distilled structural branch.

These results clarify the role of each component in DISTAL. The explicit descriptors provide a stable chemical basis. Structural distillation contributes the strongest standalone learned prior. Self-supervised compositional pretraining adds latent information that becomes most useful in the fused representation. The overall performance of DISTAL therefore arises from feature complementarity rather than dominance by a single representation. This finding supports the central design of the framework: composition-only inference can benefit from both descriptor-level chemistry and structure-aware knowledge transfer, especially when these signals are integrated within a shared downstream predictor.

## Task-wise comparison with composition-only baselines

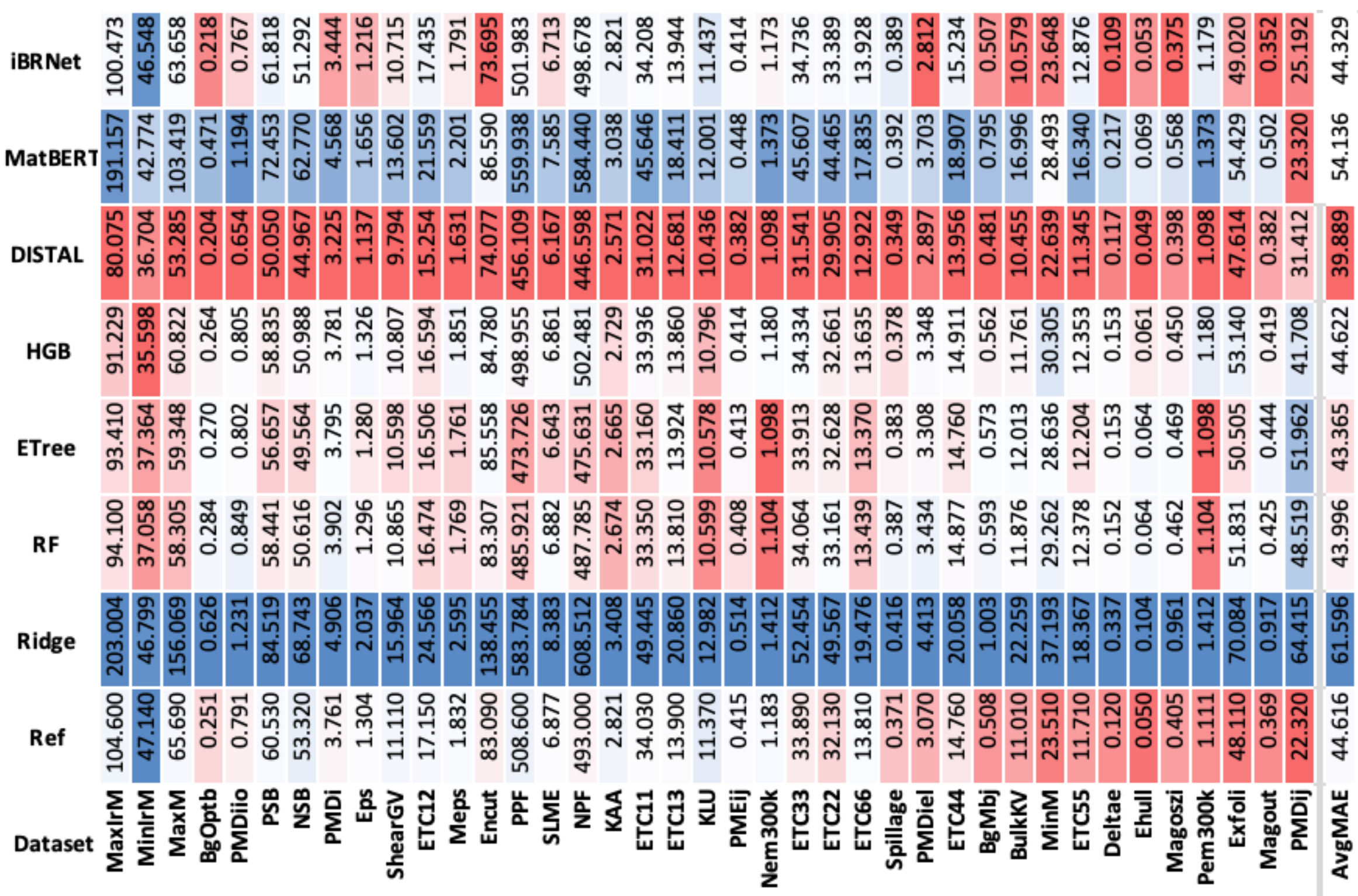

**Figure 7. Task-wise MAE comparison among the reference benchmark, external composition-only neural baselines, x-only baselines, and the final DISTAL model across 39 materials property prediction tasks.** The heatmap reports MAE values for the reference benchmark (Ref), $x$-only baseline models, MatBERT-Formula, iBRNet-EF, and the final DISTAL model. The x-only baselines were trained using only the 145-dimensional composition-derived descriptor vector $x$, including Ridge regression (Ridge), random forest (RF), extra trees (ETree), and histogram gradient boosting (HGB). MatBERT-Formula and iBRNet-EF denote additional composition-only neural baselines evaluated under the same fixed-split protocol. DISTAL denotes the final $z + x + s$ configuration. For all models trained or re-evaluated in this study, the reported MAE values were computed on the fixed external test split by averaging predictions from the 10 fold-specific models. Within each dataset column, cell backgrounds follow a blue-to-red gradient from the lowest to the highest MAE.

To place the final model in a broader task-wise context, we further compare the $z + x + s$ configuration with the reference 39-task benchmark of Gupta et al. (Gupta, et al., 2021), four $x$-only baselines trained in this study, and two additional composition-only neural baselines. The $x$-only baselines use the same 145-dimensional composition-derived descriptor vector $x$, but do not use self-supervised compositional features or distilled structural features. For each task, we evaluate the MAE values among Ridge regression, random forest, extra trees, and histogram gradient boosting. In addition, we evaluated MatBERT-Formula, inspired by LLM4Mat-Bench (Niyongabo Rubungo, et al., 2025), and iBRNet-EF, based on the elemental-fraction representation used in iBRNet (Gupta, et al., 2024). As shown in Figure 7, DISTAL improves over the reference benchmark on 37 of the 39 tasks and achieves the lowest MAE in most task-wise comparisons.

The task-wise comparison in Figure 7 also shows that the advantage of DISTAL is not confined to a small subset of datasets. Compared with the reference benchmark, the final DISTAL model improves most tasks across different property families, including band-gap-related, elastic, dielectric, thermoelectric, and piezoelectric targets. Compared with the best $x$-only baselines, DISTAL also remains competitive across the majority of tasks, indicating that the fused representation provides information beyond what can be recovered by conventional regressors trained on the raw descriptor vector alone. The comparison with MatBERT-Formula and iBRNet-EF further places DISTAL against recent composition-only neural baselines. This comparison strengthens the central interpretation of DISTAL: explicit compositional descriptors, self-supervised compositional features, and distilled structural priors each contribute distinct information, and their integration yields a stronger composition-only predictor.

The comparison also defines the current boundary of the framework. The final $z + x + s$ model did not improve over the reference benchmark on two tasks: magmom_outcar and piezo_max_dij. These exceptions suggest that some properties may remain difficult to capture from composition-only inputs, even after compositional pretraining and structural distillation. This limitation is expected for targets that may depend strongly on detailed structural, electronic, or symmetry-related information. The result does not weaken the overall benchmark pattern. Instead, it identifies where richer supervision, task-specific modeling, or direct structure-aware inputs may still be needed.

Overall, the final benchmark comparison confirms that DISTAL provides a broadly effective structure-agnostic prediction framework. The model retains the practical advantage of composition-only inference, while it benefits from information learned through both self-supervised compositional pretraining and ALIGNN-guided structural distillation.

## Conclusions

In this work, we presented DISTAL, a dual-prior framework for structure-agnostic materials property prediction. DISTAL combines self-supervised compositional pretraining with structure-aware knowledge distillation. The framework addresses a central tension in low-data materials informatics. Composition-only models are well suited for early-stage screening, but structure-aware models can use richer physical priors. DISTAL bridges this gap by retaining composition-only inference and by incorporating transferable compositional regularities and distilled structural information within a unified prediction pipeline.

Using a large virtual composition space and a 145-dimensional composition-derived descriptor representation, we first showed that raw compositional descriptors provide a strong baseline across the 39-task benchmark. We then showed that self-supervised compositional pretraining improves downstream transfer in a representation-dependent manner. In parallel, ALIGNN-guided distillation produced the strongest standalone learned branch among the evaluated feature types. Most importantly, the multimodal analysis showed that these components are complementary rather than redundant. The best overall performance was obtained by integrating explicit compositional descriptors, learned compositional features, and distilled structural features. Under the adopted evaluation protocol, the final $z + x + s$ configuration improved over the reference benchmark on 37 of 39 tasks and achieved the lowest benchmark-wide mean MAE among the evaluated feature modes.

These findings suggest that robust composition-only prediction should not rely exclusively on handcrafted descriptors, self-supervised compositional learning, or transferred structural priors. A more effective route is to combine these sources in a way that preserves their distinct strengths. In this sense, DISTAL should not be viewed as a replacement for direct structure-aware modeling. Instead, it provides a practical route for transferring part of the value of structural supervision into settings where only composition is available at inference time. This property is especially relevant to early-stage materials screening, where rapid prioritization is needed before reliable crystal structures are available.

Future work can extend DISTAL in several directions. Chemically constrained virtual-composition generation may provide a more targeted pretraining space. Richer structural teacher networks may provide more informative distillation targets. Task-aware deployment workflows may further improve practical screening under specific materials-design objectives. More broadly, this study supports dual-prior learning as a useful direction for low-data materials property prediction, where large-scale compositional pretraining and transferred structural bias can be combined within a composition-only inference framework.

## Acknowledgements

This work was supported by the Natural Science Foundation of Jilin Province YDZJ202301ZYTS288 and the Fundamental Research Funds for the Central Universities (JLU).